# Mean-field methods for a special class of Belief Networks

**Chiranjib Bhattacharyya** CBCHIRU@CSA.IISC.ERNET.IN
*Dept. of Computer Science & Automation,*
*Indian Institute of Science,*
*Bangalore,*
*Karnataka , India*
*560012*

**S. Sathiya Keerthi** MPESSK@GUPPY.MPE.NUS.EDU.SG
*Dept. of Mechanical and Production Engineering,*
*National University of Singapore,*
*Singapore*

## Abstract

The chief aim of this paper is to propose mean-field approximations for a broad class of Belief networks, of which sigmoid and noisy-or networks can be seen as special cases. The approximations are based on a powerful mean-field theory suggested by Plefka. We show that Saul, Jaakkola, and Jordan's approach is the first order approximation in Plefka's approach, via a variational derivation. The application of Plefka's theory to belief networks is not computationally tractable. To tackle this problem we propose new approximations based on Taylor series. Small scale experiments show that the proposed schemes are attractive.

## 1. Introduction

Bayesian belief networks (Pearl, 1988; Lauritzen & Spiegelhalter, 1988) are powerful graphical representations of probabilistic models. These networks are directed acyclic graphs (DAGs) whose nodes represent random variables while the links represent causal influences. This association between graph theory and probability serves as an elegant tool for handling uncertainty in real life. These networks are increasingly being used in diverse areas from image processing (Agosta, 1990) to medical diagnosis (Shwe & Others, 1991).

The usefulness of these networks relies heavily on solving the problem of inference. A large number of algorithms have been developed to efficiently compute the likelihood exactly, examples include pruning based methods (Kjaerulff, 1998), or the bounded conditioning method (Horvitz, Suermondt, & Cooper, 1989) etc. However these algorithms are slow (Jensen, Kong, & Kjaerulff, 1995) when applied to densely connected belief networks(BNs). In large networks exact algorithms are intractable, as they require summing over an exponentially large number of hidden states. A possible approach to tackle this problem is to resort to Markov Chain Monte Carlo based methods like Gibbs Sampling (Geman & Geman, 1984; Neal, 1992). This approach yields accurate results but is extremely slow in convergence.

In this paper we will focus on the mean-field theory borrowed from statistical mechanics. The intuition behind these methods is that a node is relatively insensitive to particular settings of the values of its neighbors, but rather depends on their mean-values. This observation leads to a simple and often very accurate procedures; (see Jordan, Ghahramani, Jaakkola, & Saul, 1997). This technique yields an estimate of the means of the uninstantiated nodes and also an estimate of the partition function. The main aim of this paper is to study a powerful mean-field technique due to Plefka (1982), and apply it to BNs.

Plefka initially proposed his theory in the context of spin glasses, hence it's application to Boltzmann Machines (BMs)(Ackley, Hinton, & Sejnowski, 1985) is straightforward. To develop mean-field theories for BNs on the lines of BMs, it is important that we set up a framework in which both these networks can be studied. Both BMs and BNs can be seen as different realizations of "Graphical models" (GMs). In Section 2 we review the definitions of BMs and BNs as special cases of GMs and formulate the associated probability distribution as Boltzmann Gibbs distribution.

Our main results are in Section 3. In this section Plefka's approach is derived from a variational perspective. Let $Z$ denote the partition function associated with the Boltzmann distribution. The variational perspective corresponds to introducing extra variables and deriving a convex function of these variables as an upper bound on the negative logarithm of $Z$. Tightening of this bound leads to a minimization problem for which stationarity conditions are both necessary and sufficient for global minimum. At the stationary point the bound is attained. The convex function mentioned above is computationally intractable. It is approximated and the stationarity conditions of this approximate function yields the mean-field equations. Plefka's approach provides a systematic way of tractably approximating this function to desired orders of accuracy through Taylor series. We show that the approximation obtained by Saul et al.'s approach is the same as that of the first order approximation in Plefka's approach. The extension of Plefka's approach is not direct. We propose a new activation-independent scheme based on Taylor series methods to achieve this. In Section 4 experiments on small networks were conducted. These experiments show that the obtained approximations are quite attractive.

## 2. Review of BMs and BNs

In this section we establish a framework in which both BMs and BNs can be studied. Indeed both BMs and BNs can be seen as different types of GMs, the difference being in the connectivity and the energy function used in computing the probability.

In this paper we will restrict ourselves to binary valued units. We thus define a GM to be consisting of a fixed number of stochastic units. Each unit has an associated binary random variable $S \in \{0,1\}$ (or $\{1,-1\}$). Each state $\vec{S} = \{S_1, S_2, \ldots, S_N\}$ of a GM has an associated "energy function" , which is used to define the Boltzmann Distribution:

$$P(\vec{S}) = \frac{e^{-E(\vec{S})/T}}{Z} \tag{1}$$

The denominator Z, called the partition function, is the normalization factor defined by $Z = \sum_{\vec{S}} e^{-E(\vec{S})/T}$.

GMs have units called "hidden", and "visible" units. The state vector $\vec{S}$ can be thought of as being divided into the pair of $\vec{H}$ (hidden units) and $\vec{V}$ (visible units), or $\vec{S} = \{\vec{H}, \vec{V}\}$. During operation phase, visible units are clamped, that is $\vec{V} = \vec{v}$, or $\vec{S} = \{\vec{H}, \vec{v}\}$. The distribution associated with the hidden units for the clamped phase is

$$P(\vec{H}|\vec{v}) = \frac{e^{-E(\vec{S})/T}}{Z_c(\vec{v})} \tag{2}$$

where $Z_c(\vec{v}) = \sum_{\vec{H}, \vec{V}=\vec{v}} e^{-E(\vec{S})/T}$ is used for "inferring" the values of hidden units.

When $E$ is substituted with

$$E(\vec{S}) = -0.5 \sum_{i=1}^{N} \sum_{j=1}^{N} w_{ij} S_i S_j - \sum_{i=1}^{N} h_i S_i \tag{3}$$

and also the restriction $w_{ij} = w_{ji}$, is imposed on the weights the well known BM is obtained.

In this paper we will restrict ourselves to a special class of BNs with the following energy function

$$E(\vec{S}) = -\sum_{i=1}^{N} \{S_i \ln f(M_i) + (1 - S_i) \ln(1 - f(M_i))\} \tag{4}$$

where $M_i = \sum_{j=1}^{i-1} w_{ij} S_j + h_i$, and $f$ is a function from a subset $C$ of the real line denoted by $\Re$, to the interval $(0, 1)$, i.e.,

$$f : C \subseteq \Re \rightarrow (0, 1).$$

We will assume that $f$ is an analytic function. There is also the restriction on weights $w_{ij} = 0$, if $i < j$. Sigmoid and noisy-or networks are two such special networks (Neal, 1992), with activation functions $\sigma(x) = \frac{1}{1+e^{-x}}$ and $\rho(x) = 1 - e^{-x}$ respectively. In sigmoid networks $x$ can be any real number, while in noisy-or $x$ is restricted to be non-negative; this constraint is enforced by forcing the weights and biases to be non-negative. From here on whenever we refer to BNs it would be with reference to (4).

For inference the crucial things to be evaluated are $Z_c$ and $Z$. The first item requires summing $2^H$ terms while the next term requires summing $2^N$ terms (where H is the number of hidden units and N is the total number of units). The expressions for partition functions are similar in nature both in the clamped and the unclamped phase. The only difference is that in the clamped phase the summation is over hidden units, while in the unclamped phase it is over all the units. Thus throughout the remaining paper we will talk only about evaluating $Z$.

As mentioned above computing $Z$ requires an exponential summation operation. Thus exact calculation of $Z$ is intractable. Sampling based methods offer a possible way out. But these methods are computationally very expensive. Another alternative is to look for deterministic approaches, based on mean-field theory as advocated by (Peterson & Anderson, 1987; Saul et al., 1996; Kappen & Rodriguez, 1998). The methods proposed by (Peterson & Anderson, 1987; Kappen & Rodriguez, 1998) are applicable to BMs, while Saul et al.(1996) developed a scheme which has been applied to both BNs and BMs. The next section will be devoted to Plefka's method. We intend to study it, understand its relations with existing theories and apply it to BNs.

## 3. Plefka's Method and BNs

Recently Kappen and Rodriguez(1998) introduced, to the neural network community, a powerful approximation method due to Plefka. In this section we will present Plefka's method in a different way, so that it can be extended to BNs. We will also demonstrate, later in this section that Plefka's method is more general than SJJ approach.

### 3.1 Plefka's Method

Define a new energy function

$$\widetilde{E}(\vec{\theta},\gamma) = \gamma E/T - \sum_{i=1}^{N} \theta_i S_i \tag{5}$$

(Since the dependence of various functions on $\vec{S}$, $T$ and $\vec{w}$ is quite obvious, in what follows we will not mention these as functional arguments. We do this so that the dependence on other variables such as $\gamma$ and $\vec{\theta}$ stands out more clearly.) The corresponding partition function and probability distribution are:

$$\widetilde{Z} = \sum_{\vec{S}} e^{-\widetilde{E}} \tag{6}$$

$$\widetilde{p}_\gamma = \frac{e^{-\widetilde{E}}}{\widetilde{Z}} \tag{7}$$

We treat $\theta_i$ as external variables. This helps us in extending the approach to derive mean-field theories for BNs. Our approach differs from that of (Kappen & Rodriguez, 1998), which identifies $\theta_i$ with $h_i$, the bias variables in BM, see equation (3).

The motivation for bringing in $\gamma$ and $\vec{\theta}$ and defining the above functions is as follows. If the energy function is of the linear form, $\sum_i \theta_i S_i$, then the corresponding probability distribution is factorial, (of course one can use other functions and still the probability distribution can be factorial), and computations can be done tractably. The parameter $\gamma$ can be thought of as a homotopy parameter that smoothly brings in the true energy function into picture as $\gamma$ is increased from 0. To get the original energy function from $\widetilde{E}$ we need to set

$$\theta_i = 0 \; \forall \; i \in \{1, 2, \cdots N\} \quad , \quad \gamma = 1 \tag{8}$$

For convenient manipulation, it is useful to define the function,

$$C(\vec{\theta},\gamma) = -\ln \widetilde{Z} + \sum_i \theta_i u_i \tag{9}$$

where

$$u_i = \langle S_i \rangle_{\widetilde{p}_\gamma} = \frac{\partial \ln \widetilde{Z}}{\partial \theta_i} \tag{10}$$

Since $\vec{u}$, the mean vector, is physically more meaningful than $\vec{\theta}$, it is appropriate to consider it as free and treat $\vec{\theta}$ as being dependent on $\vec{u}$. To facilitate such a treatment the following assumption is made.

**Invertibility Assumption .** For each fixed $\vec{u}$ and $\gamma$, (10) can be uniquely solved for $\vec{\theta}$.

Let $\theta_i = \kappa_i(\vec{u}, \gamma)$ denote the inverse of (10). When the inverse function is used, $C$ transforms to a function of $\vec{u}$ and $\gamma$:

$$G(\vec{u}, \gamma) = C(\vec{\theta}, \gamma)|_{\vec{\theta}=\vec{\kappa}(\vec{u},\gamma)} \tag{11}$$

Thus

$$G(\vec{u}, \gamma) = -\ln \widetilde{Z} + \sum_i \theta_i u_i \tag{12}$$

where $\theta_i$ is considered to be a function of $\vec{u}$,and $\gamma$. Note that

$$\frac{\partial G}{\partial u_i} = \sum_j \frac{\partial C}{\partial \theta_j}\frac{\partial \kappa_j}{\partial u_i} = \sum_j [-\frac{\partial \ln \widetilde{Z}}{\partial \theta_j} + u_j + \sum_k \theta_k \frac{\partial u_k}{\partial \theta_j}]\frac{\partial \kappa_j}{\partial u_i} = \theta_i \tag{13}$$

The relationship between $\vec{\theta}$ and $\vec{u}$, more precisely equations (10), (13), and the invertibility assumption, is well known and is called the Legendre Transformation (Rockafeller, 1972). Central to this transformation is the Invertibility assumption. The validity of this assumption at $\gamma = 1$ is important for the techniques discussed in the paper.

In the above we have made use of the fact that $\frac{\partial \vec{u}}{\partial \vec{\theta}}$ is the inverse of $\frac{\partial \vec{\kappa}}{\partial \vec{u}}$. Hence the $\theta_i = 0 \; \forall i$, requirement mentioned in (8) translates to

$$\frac{\partial G}{\partial u_i} = \theta_i = 0 \;\; \forall \; i \tag{14}$$

Clearly, when this constraint is enforced and $\gamma$ is carried to 1, we get $G = -\ln Z$.

The inversion of (10) is intractable for $\gamma \neq 0$, which makes it impossible to write an algebraic expression for $\kappa$ and hence G. To circumvent this problem, an approximate description of G is built by using the fact that the inversion of (10) is quite straightforward at $\gamma = 0$, and expanding $G(\vec{u}, \gamma)$ around $\gamma = 0$ using Taylor Series.

At $\gamma = 0$,

$$\widetilde{p}_0 = \prod_i \frac{e^{\theta_i S_i}}{1 + e^{\theta_i}} \tag{15}$$

and equation (10) becomes

$$u_i = \frac{e^{\theta_i}}{1 + e^{\theta_i}}$$

The inversion of this equation can be carried out explicitly to obtain $\theta_i = \ln \frac{u_i}{1-u_i}$ and

$$G(\vec{u}, 0) = \sum_i [u_i \ln u_i + (1 - u_i) \ln(1 - u_i)]$$

Also, the distribution in (15) can be expressed as a function of $\vec{u}$, in factorial form

$$\widetilde{p}_0 = \prod_i u_i^{S_i}(1 - u_i)^{1-S_i} \tag{16}$$

The truncated Taylor series approximation of $G$ is

$$\widetilde{G}_M(\vec{u},\gamma) = G(\vec{u},0) + \sum_{k=1}^{M} \frac{\gamma^k}{k!} \frac{\partial^k G}{\partial \gamma^k}\bigg|_{\gamma=0} \tag{17}$$

$\widetilde{G}_M$ is a function of $\vec{u}$ and $\gamma$, and can be viewed as an Mth order approximation of G. Let us now consider the requirements mentioned in (8). Setting $\gamma = 1$ is straightforward to implement. The requirement, $\theta_i = 0$, for all i, can be approximately enforced using (14) as

$$\theta_i = \frac{\partial G}{\partial u_i} \approx \frac{\partial \widetilde{G}_M}{\partial u_i} = 0 \ \ \forall \ i \tag{18}$$

These equations are used to set up the fixed point equations. Henceforth we will refer them as the mean-field equations.

The feasibility of the scheme lies in computing the partial derivatives of G with respect to $\gamma$ at $\gamma = 0$. In this paper we will restrict ourselves to $M = 2$. The relevant expressions (Plefka, 1982; Bhattacharyya & Keerthi, 1999b) are

$$\frac{\partial G}{\partial \gamma}\bigg|_{\gamma=0} = \langle E \rangle_{\tilde{p}_0} \tag{19}$$

$$\frac{\partial^2 G}{\partial \gamma^2}\bigg|_{\gamma=0} = -\langle (E - \langle E \rangle)^2 \rangle_{\tilde{p}_0} + \sum_{i=1}^{N} \frac{\langle (E - \langle E \rangle)(S_i - u_i) \rangle^2_{\tilde{p}_0}}{u_i(1-u_i)} \tag{20}$$

Expressions for other derivatives can also be derived in a straightforward way. For BM these derivatives are tractable. For BNs computing these terms is problematic, and so one is forced to make further approximations. New approximation schemes that we have devised will be discussed later.

To evaluate the quantities required for learning one has to compute $\frac{\partial \ln Z}{\partial w_{ij}}$. Let us consider the function

$$f(\vec{w}) = -G(\vec{u}^*, 1) = \ln Z$$

where $\vec{u}^*$ is no longer a free vector, but it is considered to be a function of $\vec{w}$ given by (14). Note that

$$\frac{\partial \ln Z}{\partial w_{ij}} = \frac{\partial f}{\partial w_{ij}} = -\left( \frac{\partial G(\vec{u}^*, 1)}{\partial w_{ij}} + \sum_k \frac{\partial G}{\partial u_k^*} \frac{\partial u_k^*}{\partial w_{ij}} \right)$$

Using (14) we get

$$\frac{\partial \ln Z}{\partial w_{ij}} = -\frac{\partial G(\vec{u}^*, 1)}{\partial w_{ij}}$$

The right hand side is intractable, but can be approximated using $\widetilde{G}_M$ at $\gamma = 1$. The derivatives required for learning can thus be computed by the approximation:

$$\frac{\partial \ln Z}{\partial w_{ij}} = -\frac{\partial G(\vec{u}^*, 1)}{\partial w_{ij}} \approx -\frac{\partial G_M(\vec{\tilde{u}}, 1)}{\partial w_{ij}}$$

where $\vec{\tilde{u}}$ is a solution of the mean-field equations, (18).

### 3.2 Saul, Jaakkola and Jordan (SJJ) Approach and Plefka's Method

Saul et al.(1996) pioneered the application of mean-field theory to BNs. In this subsection we will show that their approach can be seen as a first order approximation of Plefka's method. To establish this we present a new variational derivation of Plefka's method.

#### 3.2.1 Saul, Jaakkola and Jordan (SJJ) Approach

In this section we review Saul et al.'s work. They adopt a variational approach: form a function (with extra variables) which is a lower bound for $\ln Z$ and choose the variables so as to maximize the function. An approximate distribution, $q(\vec{S}, \vec{u})$, is chosen with a parameter vector $\vec{u}$. The parameter $\vec{u}$, is tuned so that $q(\vec{S}, \vec{u})$ is as close as possible to $p(\vec{S})$. As a measure of closeness between two distributions Kullback Leibler distance is used. It is defined by

$$D(q,p) = \sum_{\vec{S}} q(\vec{S}, \vec{u}) \ln \frac{q(\vec{S}, \vec{u})}{p(\vec{S})} \tag{21}$$

The $\vec{u}$ is chosen so that $D(q,p)$ is minimized. $q(\vec{S}, \vec{u})$ is chosen to be factorial, i.e.

$$q(\vec{S}, \vec{u}) = \prod_{i=1}^{N} u_i^{S_i} (1-u_i)^{1-S_i} \tag{22}$$

When $p$ is the Boltzmann distribution the Kullback Leibler distance takes the form

$$D(q,p) = \ln Z + L_s(\vec{u}) \tag{23}$$

where

$$L_s(\vec{u}) = \frac{1}{T} \langle E(\vec{s}) \rangle_{q(\vec{s})} + \sum_{i=1}^{N} (u_i \ln u_i + (1-u_i)\ln(1-u_i)) \tag{24}$$

Using the fact that $D(q,p) \geq 0$ we get a lower bound on $\ln Z$:

$$\ln Z \geq -L_s \tag{25}$$

$L_s$ is minimized with respect to $\vec{u}$ to get an upper bound on $-\ln Z$. The vector $\vec{u}$ is determined by solving the following set of equations.

$$\frac{\partial L_s}{\partial u_i} = 0 \tag{26}$$

Though this approximation is well known in statistical physics literature (Parisi, 1988) as the naive mean-field theory, Saul et al.'s application of this theory to BNs is new and interesting, hence we will refer the above approach as the SJJ approach. A look at (24) shows that this approach has computational feasibility only when $\langle E \rangle_q$ can be expressed as a function of $\vec{u}$. For BNs, the calculation of $\langle E(\vec{S}) \rangle_q$is intractable, even when $q$ is factorial. Saul et al. suggest exploiting activation-dependent convexity properties to develop further approximations to $L_s$. Thus one has to tailor the SJJ approach for various activation functions. An approximation, $L_{sapprox}$, is developed keeping two things in mind: (1) it is close enough to $L_s$; and (2) it is a tractable function of $\vec{u}$. Then (26) is approximated by

$$\frac{\partial L_{sapprox}}{\partial u_i} = 0 \tag{27}$$

Even when $\vec{u}$ is chosen to minimize $L_s$, the equality in (25) is attained if and only if $p$ is factorial. This follows from the fact that $D(q,p)=0$ if and only if $p=q$, and the fact that $q$ is chosen to be factorial. Thus SJJ approach is inadequate for obtaining an arbitrarily close approximation to $\ln Z$, for a general distribution, $p$. One way of overcoming this drawback, is to treat a small number of variables exactly and approximate the rest; see (Jaakkola & Jordan, 1999; Saul & Jordan, 1996). In this paper we study Plefka's approach which uses Taylor series to give a systematic way of building an arbitrarily close approximation to $\ln Z$. In the subsection we will rederive Plefka's method from a variational perspective, and show that SJJ approach is a special case of Plefka's approach.

### 3.2.2 Plefka's Method: A Variational Perspective

In this section we rederive Plefka's method from a variational perspective. The inequality on $-\ln Z$ derived in the process, and the establishment of convexity of the approximation function with respect to the variational variables are new contributions made in this paper. Furthermore, we also demonstrate that the SJJ approach is a special case of Plefka's method.

As in Section 3.3 let $\gamma$ be a real parameter that takes values from 0 to 1. Let us define a $\gamma$ dependent partition and distribution function,

$$Z_\gamma = \sum_{\vec{S}} e^{-\gamma E(\vec{S})/T} \tag{28}$$

$$p_\gamma = \frac{e^{-\gamma E(\vec{S})/T}}{Z_\gamma} \tag{29}$$

Note that $Z_1 = Z$ and $p_1 = p$. Introducing an external real vector $\vec{\theta}$, let us rewrite (28) as

$$Z_\gamma = \sum_{\vec{S}} \frac{e^{-\gamma \frac{E}{T} + \sum_i \theta_i S_i}}{\tilde{Z}} e^{-\sum_i \theta_i S_i} \tilde{Z} \tag{30}$$

where $\tilde{Z}$ is given by (6). Using Jensen's Inequality, $\langle e^{-x} \rangle \geq e^{-\langle x \rangle}$, we get

$$Z_\gamma = \tilde{Z} \sum_{\vec{S}} \tilde{p}_\gamma e^{-\sum_i \theta_i S_i} \geq \tilde{Z} e^{-\sum_i \theta_i u_i} \tag{31}$$

Taking logarithms on both sides of (31), we obtain

$$\log Z_\gamma \geq \log \tilde{Z} - \sum_i \theta_i u_i \tag{32}$$

Note that the right hand side function is nothing but $-C(\vec{\theta}, \gamma)$ as defined in (9). It is also worth noting that the motivation for defining the $C$ function comes much more naturally here.

If the invertibility assumption holds then we can use $\vec{u}$ as the independent vector (with $\vec{\theta}$ dependent on $\vec{u}$) and rewrite (32) as

$$-\ln Z_\gamma \leq G(\vec{u}, \gamma) \tag{33}$$

where $G$ is as defined in (11). This gives a variational feel to Plefka's method: treat $\vec{u}$ as external variable vector and choose it to minimize $G$. It can be further proved that $G$ is a convex function in $\vec{u}$. To see this define the Hessian of $G$ as $\mathbf{H}_{ij} = \frac{\partial^2 G}{\partial u_i \partial u_j}$. Differentiation of (10) with respect to $\vec{u}$, yields

$$\mathbf{I} = \mathbf{BH} \tag{34}$$

where

$$\mathbf{B}_{ij} = [\langle S_i S_j \rangle - \langle S_i \rangle \langle S_j \rangle] \tag{35}$$

Being a covariance matrix, $B$ is positive semidefinite. Equation (34) ensures that both $B$ and $H$ are non-singular. Hence $H$ is positive definite implying that $G$ is convex.

As discussed in Section 3.3, the difficulty in inverting (10) for $\gamma \neq 0$ remains and is solved by the Taylor series approach as discussed in that section. If we use the first order approximation the SJJ approach is obtained. For $M = 1$, we have

$$\widetilde{G}_1(\vec{u}, \gamma) = G(\vec{u}, 0) + \frac{\gamma}{T} \langle E \rangle_{\tilde{p}_0} \tag{36}$$

where

$$G(\vec{u}, 0) = \sum_{i=1}^{N} (u_i \ln u_i + (1 - u_i) \ln(1 - u_i)).$$

In fact $\tilde{G}_1$ overestimates the $G$ function

$$-\log Z_\gamma \leq G(\vec{u}, \gamma) \leq \tilde{G}_1(\vec{u}, \gamma). \tag{37}$$

To see this (also see (Bhattacharyya & Keerthi, 1999a)) note that equation (33) can also be interpreted in terms of the divergence between the two distributions, $p_\gamma$ and $\widetilde{p}_\gamma$,that is

$$D(\widetilde{p}_\gamma, p_\gamma) = \sum_{\vec{S}} \widetilde{p}_\gamma \ln \frac{\widetilde{p}_\gamma}{p_\gamma} = \ln Z_\gamma + G(\vec{u}, \gamma) \tag{38}$$

Also noting that

$$D(\widetilde{p}_0, p_\gamma) = \sum_{\vec{S}} \widetilde{p}_0 \ln \frac{\widetilde{p}_0}{p_\gamma} = \ln Z_\gamma + \tilde{G}_1(\vec{u}, \gamma) \tag{39}$$

$$D(\widetilde{p}_0, \widetilde{p}_\gamma) = \sum_{\vec{S}} \widetilde{p}_0 \ln \frac{\widetilde{p}_0}{\widetilde{p}_\gamma} = \tilde{G}_1(\vec{u}, \gamma) - G(\vec{u}, \gamma) \tag{40}$$

and using the fact that the divergence $D$ is always non-negative (37) is obtained.

At $\gamma = 1$

$$\widetilde{G}_1(\vec{u}, 1) = L_s(\vec{u}) \tag{41}$$

where $L_s$ is the objective function obtained by SJJ approach; see (24). Note that at $\gamma = 1$, (37) establishes the inequality (25) in the SJJ approach. It is thus clearly established that SJJ approach is a first order approximation to Plefka's approach. Incidentally Barber and van de Laar (1999) also rederived the SJJ approach by using a cumulant expansion.

Using (38), (39), (40), an alternate information geometric derivation of Plefka's method can be constructed. The variational derivation presented here also helps in establishing links with other refinements like TAP and linear response correction. For a detailed discussion of this points refer to (Bhattacharyya & Keerthi, 2000, 1999b).

### 3.3 Mean-field Approximations for BNs

In this section, mean-field theory for BNs is developed using Plefka's method discussed in the previous sections. For BNs each $S_i$ is influenced by $\sum_{j=1}^{i-1} w_{ij}S_j + h_j$, which can be viewed as fields. The mean-field approximation then suggests that these probabilistic fields may be replaced by their mean values, that is $\sum_{j=1}^{i-1} w_{ij}u_j + h_i$. Keeping this in mind, Plefka's method is adapted to develop mean-field schemes for BNs. In Section 3.2 we suggest an approximation method which can be used to compute all the quantities required for implementing Plefka's approach. Our approach is quite general and does not depend on the form of activation function. For other activation independent approaches regarding the application of mean-field theory to BNs see (Haft, Hofmann, & Tresp, 1999; Kearns & Saul, 1998). Since for belief network operation $T$ is set to 1, we will drop $T$ from all further equations.

#### 3.3.1 A New Scheme Based on Plefka's Method

Plefka's method, as presented in Section 4, is not directly useful for BNs, because of the intractability of the partial derivatives at $\gamma = 0$. To overcome this problem, we suggest a method based on Taylor series expansion. This is a very general method and is not dependent on the activation function. This method enables calculation of all the necessary terms required for extending Plefka's method for BNs.

Let us define a new energy function

$$\widehat{E}(\beta, \vec{S}, \vec{u}, \vec{w}) = -\sum_{i=1}^{N} \{S_i \ln f(\widehat{M}_i(\beta)) + (1 - S_i) \ln(1 - f(\widehat{M}_i(\beta))\} \tag{42}$$

where $0 \leq \beta \leq 1$,

$$\widehat{M}_i(\beta) = \sum_{j=1}^{i-1} w_{ij}\beta(S_j - u_j) + \overline{M_i}$$

and

$$\overline{M_i} = \sum_{j=1}^{i-1} w_{ij}u_j + h_i.$$

Since $\beta$ is the important parameter, $\widehat{E}(\beta, \vec{S}, \vec{u}, \vec{w})$ will be referred to as $\widehat{E}(\beta)$ so as to avoid notational clumsiness. Note that

$$\widehat{E}(0) = -\sum_{i=1}^{N} \{S_i \ln f(\overline{M}_i) + (1 - S_i) \ln(1 - f(\overline{M}_i)\}$$

and $\widehat{E}(1) = E$. We use a Taylor series approximation of $\widehat{E}(\beta)$ with respect to $\beta$. Let us define

$$\widehat{E}_C(\beta) = \widehat{E}(0) + \sum_{k=1}^{C} \frac{\beta^k}{k!} \frac{\partial^k \widehat{E}}{\partial \beta^k}\bigg|_{\beta=0}. \tag{43}$$

If $\widehat{E}_C$ approximates $\widehat{E}$, then we can write

$$E = \widehat{E}(1) \approx \widehat{E}_C(1). \tag{44}$$

Let us now define the following function

$$A(\gamma, \beta, \vec{u}) = -\ln \sum_{\vec{s}} e^{-\gamma \widehat{E} + \sum_i \theta_i S_i} + \sum_i \theta_i u_i \tag{45}$$

The $\theta_i$ are assumed to be functions of $\vec{u}, \beta, \gamma$, which are obtained by inverting the following equations:

$$u_k = \sum_{\vec{S}} S_k p_{\gamma\beta} \;\; \forall k \tag{46}$$

where

$$p_{\gamma\beta} = \frac{e^{-\gamma \widehat{E} + \sum_i \theta_i S_i}}{\sum_{\vec{s}} e^{-\gamma \widehat{E} + \sum_i \theta_i S_i}} \tag{47}$$

By replacing $\widehat{E}$ by $\widehat{E}_C$ in (45) we obtain $A_C$

$$A_C(\gamma, \beta, \vec{u}) = -\ln \sum_{\vec{s}} e^{-\gamma \widehat{E}_C + \sum_i \theta_i S_i} + \sum_i \theta_i u_i \tag{48}$$

where the definition of $\vec{u}$ is obtained by replacing $\widehat{E}$ by $\widehat{E}_C$. In view of (44) one can consider $A_C$ as an approximation to $A$. This observation suggests an approximation to $G$.

$$G(\gamma, \vec{u}) = A(\gamma, 1, \vec{u}) \approx A_C(\gamma, 1, \vec{u}) \tag{49}$$

Then the mean-field equations can be stated as

$$\theta_i = \frac{\partial G}{\partial u_i} = \frac{\partial A}{\partial u_i}\bigg|_{\beta=1} \approx \frac{\partial A_C}{\partial u_i}\bigg|_{\beta=1} = 0 \tag{50}$$

The required terms needed in the Taylor expansion of $G$ in $\gamma$ can be approximated by

$$G(\vec{u}, 0) = A(0, 1, \vec{u}) = A_C(0, 1, \vec{u})$$

$$\frac{\partial^k G}{\partial \gamma^k}\bigg|_{\gamma=0} = \frac{\partial^k A}{\partial \gamma^k}\bigg|_{\gamma=0,\beta=1} \approx \frac{\partial^k A_C}{\partial \gamma^k}\bigg|_{\gamma=0,\beta=1}$$

The biggest advantage in working with $A_C$ rather than $G$ is that the partial derivatives of $A_C$ with respect to $\gamma$ at $\gamma = 0$ and $\beta = 1$ can be expressed as functions of $\vec{u}$. We define

$$\widehat{G}_{MC}(\vec{u}, \gamma) = A_C(\vec{u}, 0) + \sum_{k=1}^{M} \frac{\gamma^k}{k!} \frac{\partial^k A_C}{\partial \gamma^k}\bigg|_{\gamma=0,\beta=1} \tag{51}$$

In light of the above discussion one can consider

$$\widetilde{G}_M \approx \widehat{G}_{MC} \tag{52}$$

and hence the mean-field equations can be stated as

$$\theta_i = \frac{\partial G}{\partial u_i} \approx \frac{\partial G_M}{\partial u_i} \approx \frac{\partial \widehat{G}_{MC}}{\partial u_i} = 0 \tag{53}$$

To develop a feel for the approximations let us consider the case where $M = 1$, the first order case. Recall that at $\gamma = 0$, the distribution is given by

$$\tilde{p}_0 = \prod_i u_i^{S_i}(1-u_i)^{(1-S_i)}.$$

$$\left.\frac{\partial A_C}{\partial \gamma}\right|_{\gamma=0,\beta=1} = \langle \widehat{E}_C(1)\rangle_{\widetilde{p}_0} = \langle \widehat{E}(0)\rangle_{\widetilde{p}_0} + \sum_{k=1}^{C} \frac{1}{k!}\left\langle \left.\frac{\partial^k \widehat{E}}{\partial \beta^k}\right|_{\beta=0}\right\rangle_{\widetilde{p}_0} \tag{54}$$

All the terms can be computed as a function of $\vec{u}$. The appropriate expressions are derived in appendix C. All the above terms can be used in computing an approximation to $\widetilde{G}_1$. Using (51) we obtain

$$\widehat{G}_{1C}(\vec{u},\gamma) = G(\vec{u},0) + \gamma\left\{\langle \widehat{E}(0)\rangle_{\widetilde{p}_0} + \sum_{k=1}^{C} \frac{1}{k!}\left\langle \left.\frac{\partial^k \widehat{E}}{\partial \beta^k}\right|_{\beta=0}\right\rangle_{\widetilde{p}_0}\right\} \tag{55}$$

Thus using (52) we have

$$\widetilde{G}_1(\gamma) \approx \widehat{G}_{1C}(\gamma) \tag{56}$$

The mean-field equations are obtained by using (53) to get

$$\frac{\partial \widehat{G}_{1C}}{\partial u_i} = 0 \tag{57}$$

The fixed point equations thus obtained are

$$u_i = \sigma\left(-\left(\langle \widehat{E}(0)\rangle_{\widetilde{p}_0} + \sum_{k=1}^{C} \frac{1}{k!}\left\langle \left.\frac{\partial^k \widehat{E}}{\partial \beta^k}\right|_{\beta=0}\right\rangle_{\widetilde{p}_0}\right)\right) \tag{58}$$

Note that the scheme resulting from C=1 approach can also be obtained from a saddle point approach, (Bhattacharyya & Keerthi, 1999a). A confusion might arise regarding the interpretation of $\widetilde{G}_M$ The variational derivation might tempt one to believe that since $G$ is an upper bound to $-\ln Z$, and $\widetilde{G}_M$ is an approximation to $G$, then $\widetilde{G}_M$ is also an upper bound. Such arguments are not correct. It is difficult to establish whether $\widetilde{G}_M$ can be considered as an upper or lower bound for a given $M$ and $E$. In fact even for $M = 2$, one can at most say $\widetilde{G}_2 < \widetilde{G}_1$, (Bhattacharyya & Keerthi, 1999b), for a general $E$.

We interpret $\widetilde{G}_M$ as a close approximation to $G$ as a function of $\vec{u}$. If the approximations are close enough then the gradients $\frac{\partial G}{\partial u_i}$ must also be well approximated by $\frac{\partial \widetilde{G}_M}{\partial u_i}$ which ultimately leads to (18). A similar interpretation applies to $\widehat{G}_{MC}$.

Application of Plefka's method to any stochastic system, i.e. for any $E$, relies on the validity of the invertibility assumption which depends on the convergence of the Taylor series expansion of $G$ in $\gamma$. More importantly the radius of convergence $\rho$ has to satisfy $\rho > 1$ (Plefka, 1982). It is still an open question whether the radius of convergence for the energy function described in (4) or in (43) satisfies this condition.

## 4. Examples and Numerical Experiments

In this section we apply the various approximation schemes developed in the previous section to two different class of networks, namely the sigmoid network defined by the sigmoid activation function is $\sigma(x) = \frac{1}{1+e^{-x}}$, and the noisy-or network defined by the activation function $\rho(x) = 1 - e^{-x}$, $x > 0$. In this paper we will restrict ourselves to three different approximation schemes having the objective functions $\widehat{G}_{11}, \widehat{G}_{12}, \widehat{G}_{22}$. It is interesting to compare the approximations for sigmoidal BNs as derived in (Saul et al., 1996) with those obtained here. Of particular interest is $\widehat{G}_{12}$. The fixed point equations resulting from this objective function are

$$u_i = \sigma\left(\left[\overline{M}_i + \sum_{k=i+1}^{N} \{(u_k - \sigma(\overline{M}_k))w_{ki} + \widetilde{K}_{ki}\}\right]\right) \tag{59}$$

$$\widetilde{K}_{ki} = -0.5\sigma(\overline{M}_k)(1-\sigma(\overline{M}_k))\left\{(1-2\sigma(\overline{M}_k))w_{ki}\sum_{l=1}^{k-1} w_{kl}^2 u_l(1-u_l) + w_{ki}^2(1-2u_i)\right\} \tag{60}$$

If $k < i$ then $\widetilde{K}_{ki} = 0$. Approximation for sigmoidal BNs as derived in (Saul et al., 1996) has the following fixed point equation

$$u_i = \sigma\left(\left[\overline{M}_i + \sum_{k=i+1}^{N} \{(u_k - \xi_k)w_{ki} + K_{ki}\}\right]\right) \tag{61}$$

where again $K_{ki} = 0$ , if $k < i$; just like (60), $K_{ki}$ is also dependent on $u_l$, $l = 1, \cdots, k-1$.

The expressions for $K_{ki}$ (refer to Saul et al., 1996 for exact expressions) look very different from $\widetilde{K}_{ki}$. It may be still possible that numerically they may be very close. In fact the similarity in structure in (59) and (61) is worth noting, and experimental results show that as far as approximation of $\ln Z$ goes they yield close results. It is possible that the $\xi_k$ and $\sigma(\overline{M}_k)$ play the same role. One can refer to (Saul & Jordan, 1999) for a discussion on this point. It is a matter of future study to investigate the above relationships in detail.

### 4.1 Experimental Results

To test the approximation schemes developed in Section 3, numerical experiments were conducted. Small Networks were chosen so that $\ln Z$ can be computed by exact enumeration. For all the experiments the network topology was fixed to $2 \times 4 \times 6$; see figure 1. This choice of the network enables us to compare the results with those of Saul et al.(1996). To compare the performance of our methods with their method we repeated the experiment conducted by them for sigmoid BNs. $10,000$ networks were generated by randomly choosing weight values in $[-1, 1]$. The bottom layer units, or the visible units of each network was instantiated to zero. The likelihood, $\ln Z$, was computed by exact enumeration of all the states in the higher two layers. The approximate value of $-\ln Z$ was computed by $\widehat{G}_{MC}$; $\vec{u}$ is computed by solving the fixed point equations obtained from (53). The goodness of approximation scheme was tested by the following measure

$$\mathcal{E} = -\frac{\widehat{G}_{MC}}{\ln Z} - 1 \tag{62}$$

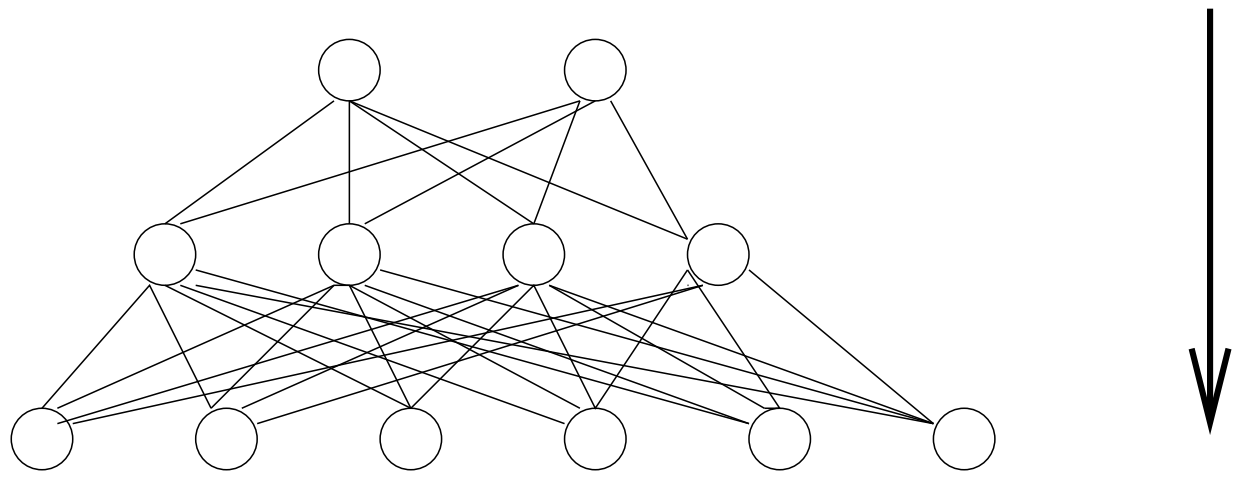

Figure 1: Three layer BN ($2 \times 4 \times 6$) with top down propagation of beliefs. The activation function was chosen to be one of sigmoid and noisy-or.

We implemented 3 approximation schemes $\widehat{G}_{11}$, $\widehat{G}_{12}$, $\widehat{G}_{22}$. Relevant expressions are worked out in Appendix B. For a proper comparison we also re-implemented the SJJ method. The goodness of approximation for the SJJ scheme is evaluated by substituting $\widehat{G}_{MC}$, in (62) by $L_{sapprox}$, mentioned in Section 3.2.1, for specific formula see (Saul et al., 1996). The results are presented in the form of histograms in Figure 2 and Figure 3. We also repeated the experiment with weights and biases taking values between -5 and 5, the results are again presented in the form of histograms in Figure 4 and Figure 5. The findings are summarized in the form of means tabulated in Table 1.

For small weights $\widehat{G}_{12}$ and the SJJ approach show close results, which was expected. But the improvement achieved by the $\widehat{G}_{22}$ scheme is remarkable, it gave a mean value of 0.0029 which compares substantially well against the mean value of 0.01139 reported by Bishop et al.. They suggest the use of mixture distributions which requires introduction of extra variational variables; more than 100 extra variational variables are needed for a 5 component mixture. This results in substantial increase in the computation costs. On the other hand the extra computational cost for $\widehat{G}_{22}$ over $\widehat{G}_{12}$ is marginal. This makes the $\widehat{G}_{22}$ scheme computationally attractive over the mixture distribution.

To study the robustness of the schemes the weight scales were increased. As the weight scales were increased degradation was noticed in all the four methods. The point to be noted is all the three methods appear to be more robust than the SJJ approach. But unlike the small weights case $\widehat{G}_{22}$ did not perform well, it is a poor approximation for large weights. The best results are obtained by $\widehat{G}_{12}$ scheme.

During the course of experimentation it was found that, for some networks with large weights, the fixed point equations converged to an order 2 fixed point; while solving the fixed point equations $\vec{u}$ oscillates between two vectors $\vec{u}_1$ and $\vec{u}_2$. To solve this problem we adapted the following strategy. As soon as the oscillations were detected we stopped the fixed point equations, and restarted it with a new point $\vec{u}_3$. This new point was chosen by searching along the line between $\vec{u}_1$ and $\vec{u}_2$, which gave a minimum value of $\mathcal{E}$. Once this was done convergence to an order 1 fixed point occurred.

| | $\langle \mathcal{E} \rangle$ small weights $[-1, 1]$ | $\langle \mathcal{E} \rangle$ large weights $[-5, 5]$ |
|---|---|---|
| $\widehat{G}_{11}$ | -0.0404 | -0.0440 |
| $\widehat{G}_{12}$ | 0.0155 | 0.0231 |
| $\widehat{G}_{22}$ | 0.0029 | -0.0456 |
| $SJJ$ | 0.0157 | 0.0962 |

Table 1: Mean of $\mathcal{E}$ for randomly generated sigmoid networks. $10,000$ networks were randomly selected by choosing the weights from the range $[-1, 1]$. The experiment was repeated by choosing the weights from a larger range $[-5, 5]$

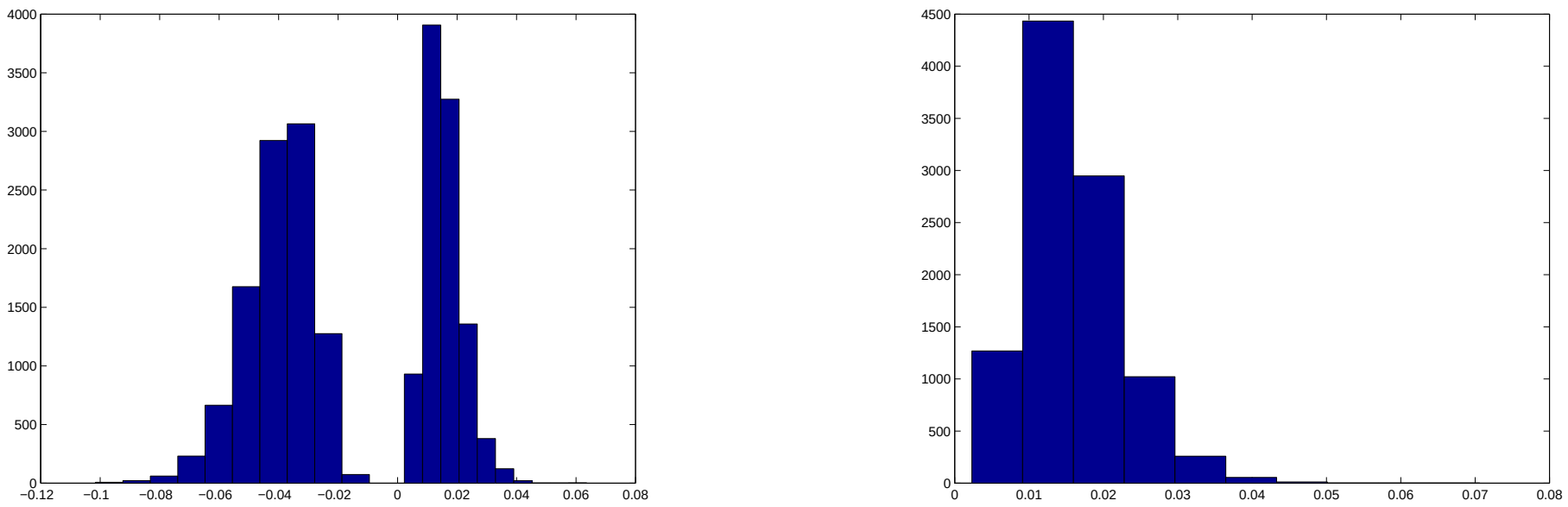

Figure 2: Histograms for $\widehat{G}_{1C}$ and SJJ scheme for small weights, taking values in $[-1, 1]$, for sigmoid networks. The plot on the left show histograms for $\mathcal{E}$ for the schemes $\widehat{G}_{11}$ and $\widehat{G}_{12}$. They did not have any overlaps and they clearly show the improvement. $\widehat{G}_{11}$, gives a mean of -0.040 while $\widehat{G}_{12}$ gives a mean of 0.0155. The plot on the right shows the histogram for the SJJ scheme. The mean is given by 0.0157.

Numerical experiments were also conducted for noisy-or networks. For the approximation schemes to work well it should be able to approximate $\ln Z$ over a range of values. This motivated the experiment described below. Noisy-or networks, whose topology is given by Figure 1 were randomly generated by choosing weights and biases randomly from 0 and 0.25. For each network $\ln Z$ was computed for all the bottom layer states. Out of all the states, two states were chosen such that $\ln Z$ is maximized and minimized respectively. The bottom layer was instantiated with each of the chosen two states, and approximations to the likelihood were then computed. This experiment was repeated for $10,000$ such networks. Again $\mathcal{E}$ is used as a measure of goodness of approximation. Figures 6 and 7 show the corresponding histograms. We repeated the experiment for a different weight range, $[0.2, 0.8]$. Figures 8 and 9 show the relevant histograms. The histograms are summarized by mean values of $\mathcal{E}$ tabulated in table 2. In this case it appears that $\widehat{G}_{22}$ is indeed a poor approximation. $\widehat{G}_{12}$ almost always gave the best results showing that even for noisy-or ac-

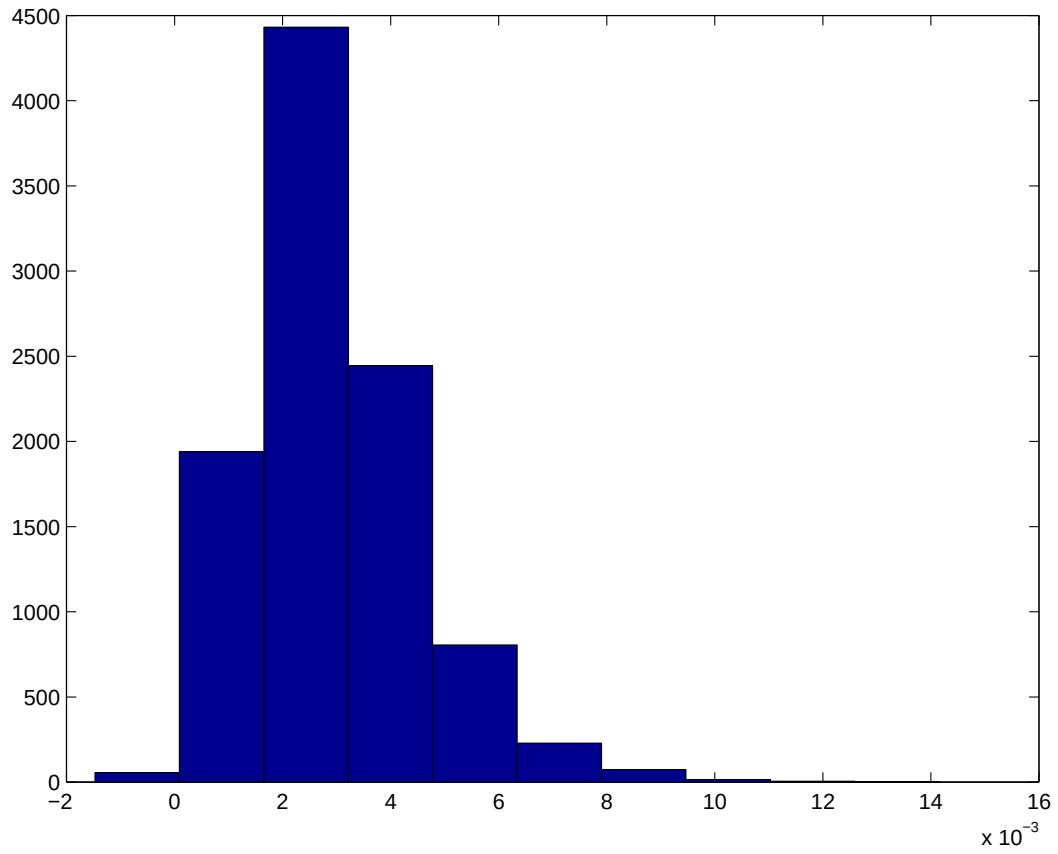

Figure 3: Histogram of $\mathcal{E}$ for $\widehat{G}_{22}$ applied to sigmoid network with small weights. The mean obtained is 0.0029. Note that $\mathcal{E}$ is scaled by $10^3$ in this figure

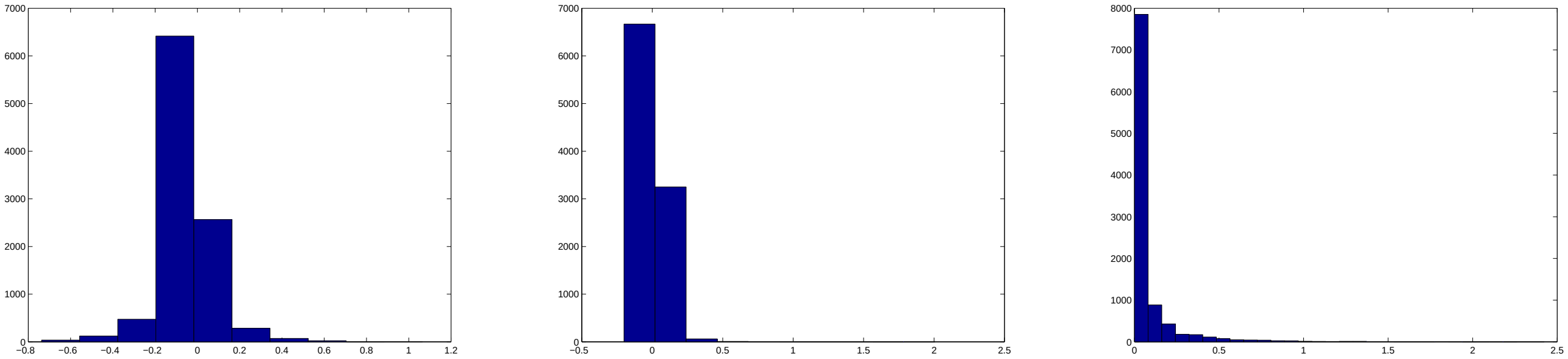

Figure 4: Histograms for the $\widehat{G}_{1C}$ and SJJ schemes for large weights, taking values in $[-5, 5]$ for sigmoid networks. The histogram on the left shows $\mathcal{E}$ for $\widehat{G}_{11}$ scheme having a mean of $-0.0440$, one at the center is for $\widehat{G}_{12}$ scheme having a mean of 0.0231, and one at the right is for SJJ scheme, having a mean of 0.0962.

| | Small weights $[0., 0.25]$ | | large weights $[0.2, 0.8]$ | |
|---|---|---|---|---|
| | $\ln Z_{max}$ case | $\ln Z_{min}$ case | $\ln Z_{max}$ case | $\ln Z_{min}$ case |
| $\widehat{G}_{11}$ | 0.001 | -0.061 | -0.156 | -0.029 |
| $\widehat{G}_{12}$ | 0.028 | 0.011 | 0.052 | 0.015 |
| $\widehat{G}_{22}$ | 0.445 | 0.320 | 0.090 | 0.211 |

Table 2: Mean of $\mathcal{E}$ for randomly generated noisy-or networks. $10,000$ networks were randomly selected by choosing the weights from the range $[0., 0.25]$. For each network the visible states with maximum $\ln Z$ and minimum $\ln Z$ were identified. The visible nodes of each network is then instantiated with the identified states, and the corresponding $\log Z$ is approximated by various schemes. The experiment was repeated by choosing the weights from a different range $[0.2, 0.8]$

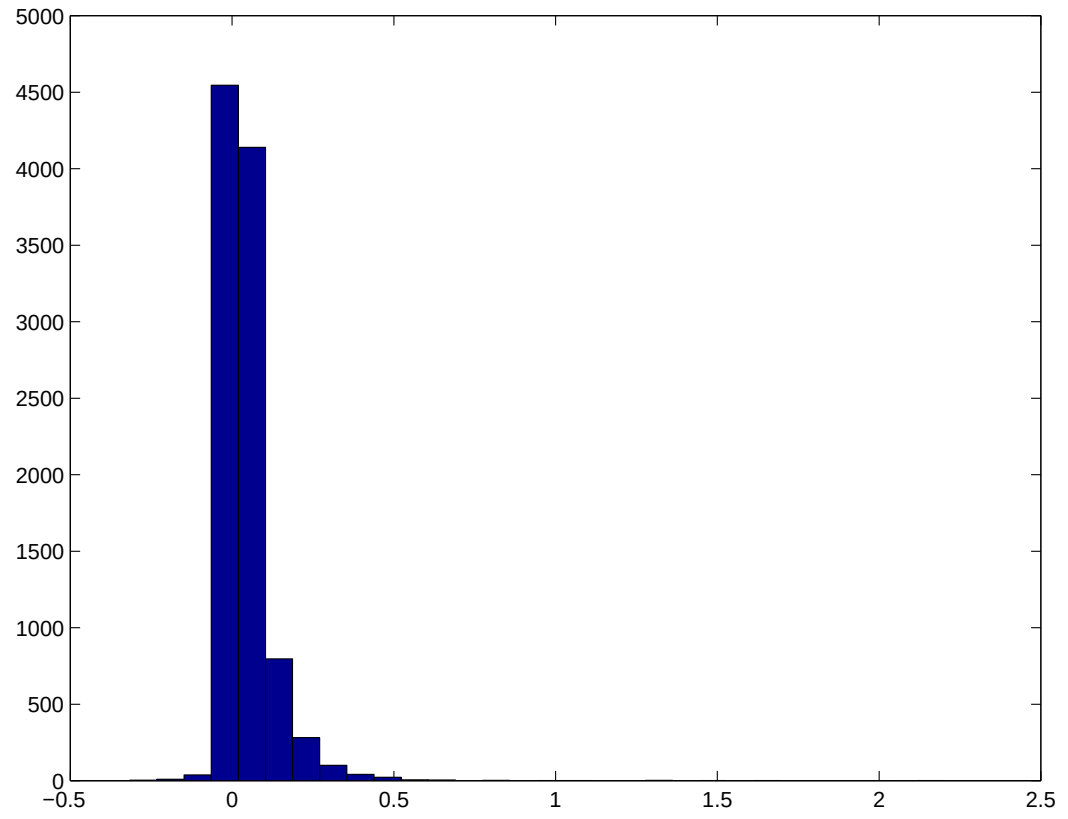

Figure 5: The histogram shows $\mathcal{E}$ for $\widehat{G}_{22}$, having a mean of $-0.0456$. The network is sigmoid with large weights

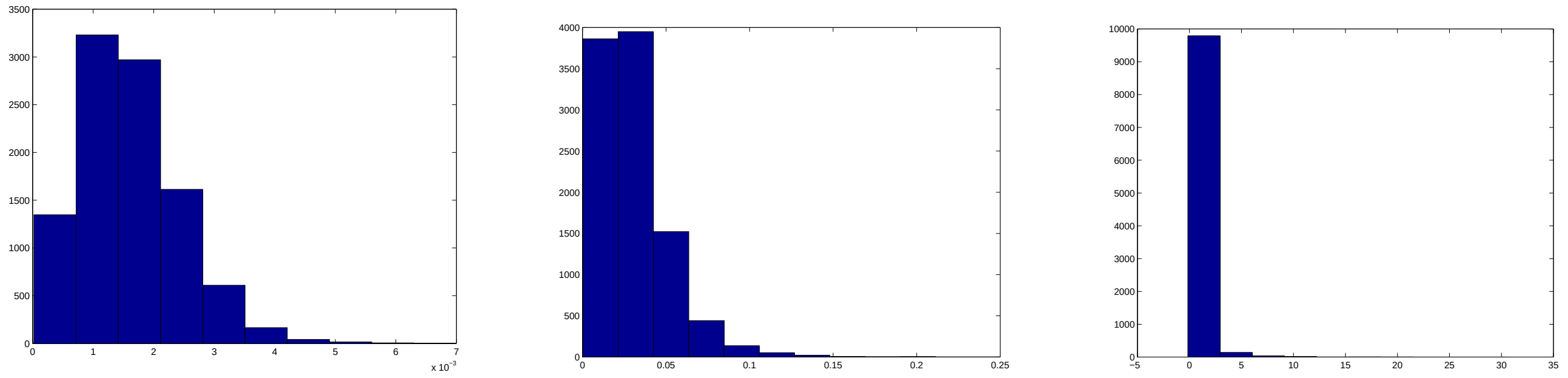

Figure 6: Histograms for visible state with maximum $\log Z$, for noisy-or networks with weights taking values in $[0, 0.25]$. The histogram on the left shows $\mathcal{E}$ for $\widehat{G}_{11}$ scheme, at the center $\widehat{G}_{12}$ scheme and that at the right $\widehat{G}_{22}$. The scheme $\widehat{G}_{11}$ gave a mean of 0.001, $\widehat{G}_{12}$ gave a mean of 0.028 and $\widehat{G}_{22}$ gave a mean of 0.445

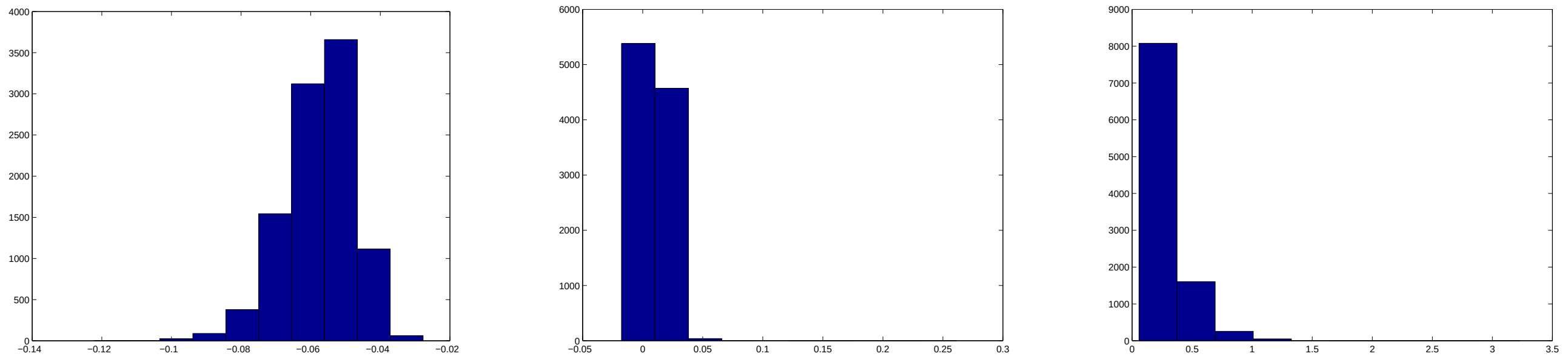

Figure 7: Histograms for visible state with minimum $\log Z$, for noisy-or networks with weights taking values in $[0, 0.25]$. The histogram on the left shows $\mathcal{E}$ for $\widehat{G}_{11}$ scheme, at the center $\widehat{G}_{12}$ scheme and that at the right $\widehat{G}_{22}$. The scheme $\widehat{G}_{11}$ gave a mean of -0.061, $\widehat{G}_{12}$ gave a mean of 0.011 and $\widehat{G}_{22}$ gave a mean of 0.320

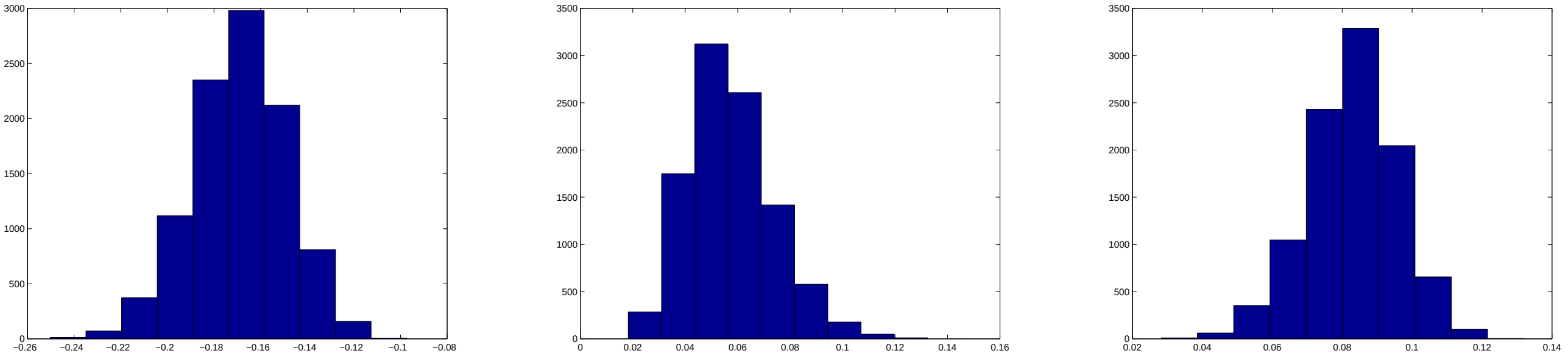

Figure 8: Histograms for visible state with maximum $\log Z$, for noisy-or networks with weights taking values in $[0.2, 0.8]$. The histogram on the left shows $\mathcal{E}$ for $\widehat{G}_{11}$ scheme, at the center $\widehat{G}_{12}$ scheme and that at the right $\widehat{G}_{22}$. The scheme $\widehat{G}_{11}$ gave a mean of -0.156, $\widehat{G}_{12}$ gave a mean of 0.052 and $\widehat{G}_{22}$ gave a mean of 0.09

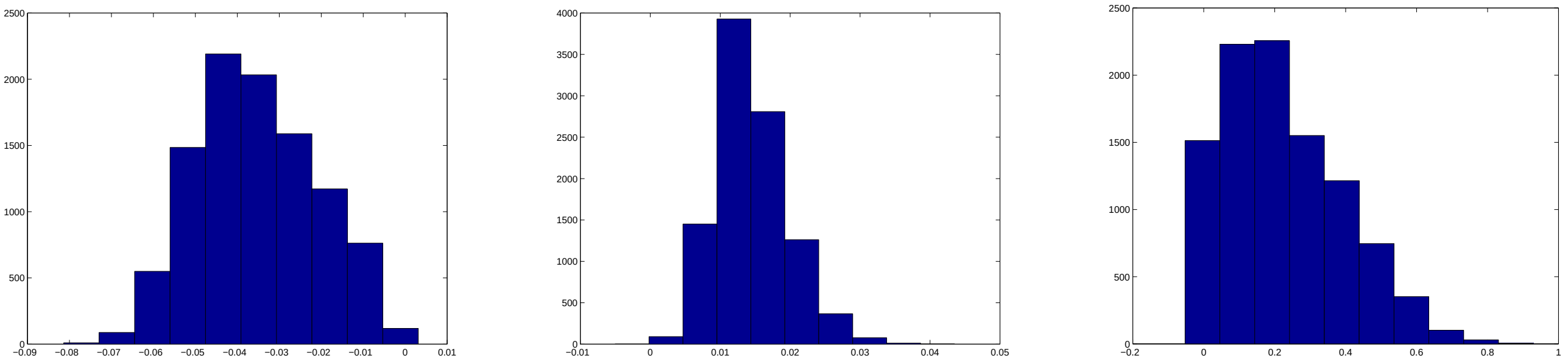

Figure 9: Histograms for visible state with minimum $\log Z$, for noisy-or networks with weights taking values in $[0.2, 0.8]$. The histogram on the left shows $\mathcal{E}$ for $\widehat{G}_{11}$ scheme, at the center $\widehat{G}_{12}$ scheme and that at the right $\widehat{G}_{22}$. The scheme $\widehat{G}_{11}$ gave a mean of -0.029, $\widehat{G}_{12}$ gave a mean of 0.015 and $\widehat{G}_{22}$ gave a mean of 0.212

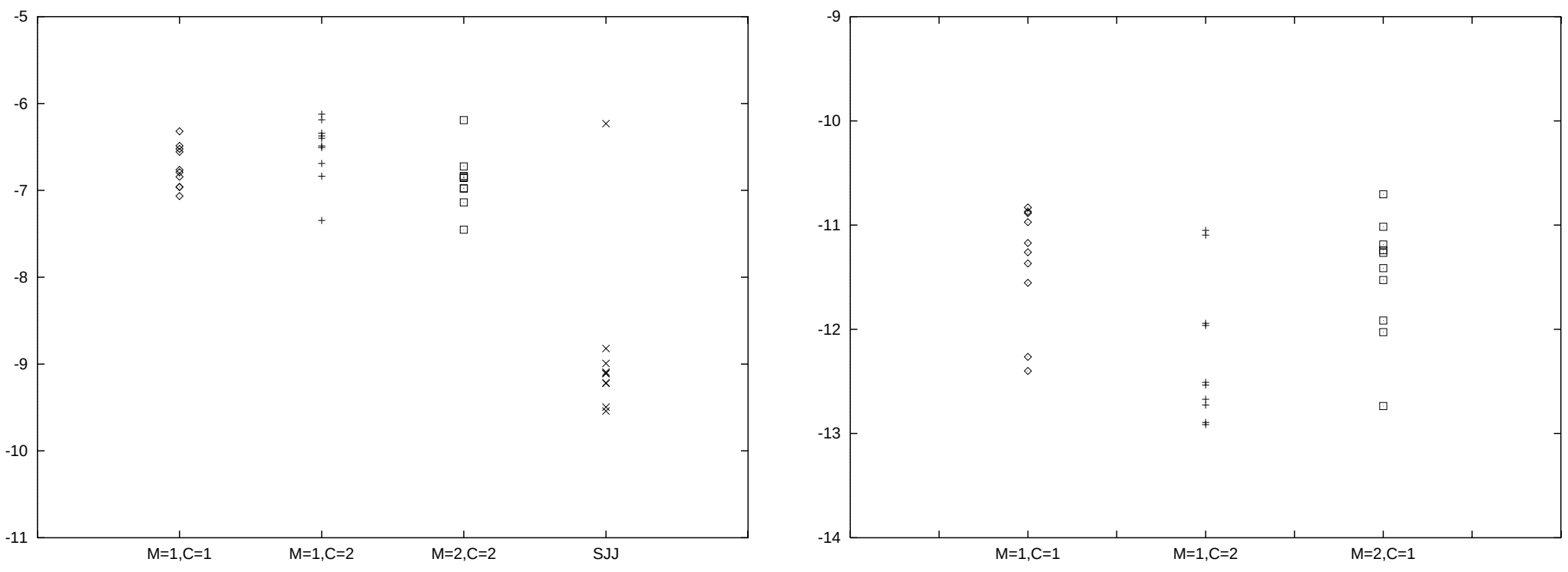

Figure 10: The plot on the left shows True log likelihood divided by the number of patterns for $\widehat{G}_{11}$, $\widehat{G}_{12}$, $\widehat{G}_{22}$ and SJJ schemes after training on sigmoid networks; the plot on the right shows the true log likelihood divided by the number of patterns for noisy-or networks

tivation function it is the best. Again it should be mentioned that for some of the networks the fixed point equations converged to an order 2 fixed point, the heuristic described before was used to solve this problem.

Of the three schemes $\widehat{G}_{12}$ is the most robust and also yields reasonably accurate results. It is outperformed only by $\widehat{G}_{22}$ in the case of sigmoid networks with low weights. Empirical evidence thus suggests that the choice of a scheme is not straightforward and depends on the activation function and also parameter values.

To study the learning capabilities of the various schemes proposed we took up a toy problem suggested by Hinton et al.(1995) involving binary images. These binary images are of size $4 \times 4$ in which each image consists of either vertical or horizontal bars with equal probability, with each location of the bar occupied with probability of 0.5. We took a $1 \times 8 \times 16$ network and tried to learn it using both the sigmoid and noisy-or activation functions. Number of patterns used was 2000, while the number of epochs was 500. The experiment was repeated for 10 different networks. For each network true likelihood was computed by exact enumeration. The SJJ method yielded lower likelihoods in almost all the cases. It thus appears that the three proposed schemes have a better learning performance than the SJJ approach. The results are summarized in figure 10.

## 5. Discussion

In this section we summarize the contributions of this paper, and identify issues for future research. The main contributions of this paper are presented in Section 3. In Section 3 Plefka's method is introduced, re-derived from an variational perspective and applied to BNs. Plefka's approach gives a systematic way of building an arbitrarily close approximation to $-\ln Z$. However it should be noted that the effort needed to evaluate higher order terms increases with the order and might be even exponential for an arbitrarily close approximation.

The variational derivation establishes that SJJ approach is a special case of Plefka's approach, thus serving as a link with the existing theory. This derivation process does not make any assumptions regarding the structure of energy function and hence it is also applicable to BNs. The validity of Plefka's method is subject to the condition, that radius of convergence should be greater than 1; see Section 3. It is still an open question whether one can prove that such condition holds for the BN energy function.

Application of Plefka's theory to BNs is not straightforward, it requires computation of some averages which are not tractable. We presented a scheme in which the BN energy function is approximated by a Taylor series, which gives a tractable approximation to the terms required for Plefka's method. Various approximation schemes depending on the degree of the Taylor series expansion are derived. Unlike the approach in (Saul et al., 1996), the schemes discussed here are simpler as they do not introduce extra variational variables; compare equations (59) and (61). Another positive aspect of these approximations is that they are general and not activity function dependent; hence they are applicable to a broad class of BNs. Of course for a given activation function it might be possible to come up with tailor-made approximations which are better than the schemes discussed here. But empirical evaluation on small scale networks show that the quality of approximations is better than those obtained from other variational methods.

The computational simplicity, robustness and generality make these schemes very attractive. Unfortunately theoretical guarantees regarding the validity of Taylor series expansion in $\beta$ are missing. This is another open issue which needs to be addressed in the near future. It would be interesting to see how these schemes perform on real world datasets. We are presently exploring the applicability of these methods on the hand-written digits data-base.

## Acknowledgments

We are extremely grateful to the various anonymous referees, whose inputs went a long way in improving the paper. We are also thankful to Dr H. J. Kappen for his suggestions.

## Appendix A: Expressions for 1st Order Approximation for BNs

In this section we present expressions for the"first order approximation" method introduced in Section 3. To compute $\langle X_i^k \rangle$, the average being taken over the factorial distribution $\prod_{j=1}^N u_j^{S_j}(1-u_j)^{1-S_j}$, we define a random variable $R_i = e^{\lambda X_i}$ where $X_i = \sum_{j=1}^{i-1} w_{ij}(S_j - u_j)$. Now

$$\langle R_i \rangle = \sum_S e^{\lambda X_i} \prod_{j=1}^N u_j^{S_j}(1-u_j)^{1-S_j} \tag{63}$$

$$\left. \frac{\partial^k \langle R_i \rangle}{\partial \lambda^k} \right|_{\lambda=0} = \langle X_i^k \rangle \tag{64}$$

$$\langle R_i \rangle = \prod_{j=1}^{i-1} \{ e^{\lambda w_{ij}(1-u_j)} u_j + e^{-\lambda w_{ij} u_j}(1-u_j) \} \tag{65}$$

$$\langle X_i \rangle = 0 \tag{66}$$

$$\langle X_i^2 \rangle = \sum_{j=1}^{i-1} w_{ij}^2 u_j (1 - u_j) \tag{67}$$

Differentiation of $\widehat{E}$ in (42) and taking average with the factorial distribution yields

$$\left\langle \frac{\partial^k \widehat{E}}{\partial \beta^k}\bigg|_{\beta=0} \right\rangle = -\sum_{i=1}^{N} g_{ki}(\vec{u}, \overline{M}_i) \langle X_i^k \rangle \tag{68}$$

where

$$g_{ki}(u_i, \overline{M}_i) = u_i \frac{\partial^k}{\partial \overline{M}_i^k} \ln f(\overline{M}_i) + (1 - u_i) \frac{\partial^k}{\partial \overline{M}_i^k} \ln(1 - f(\overline{M}_i)) \tag{69}$$

The first term $g_{1i}$ term is irrelevant beacause of (66). In our implementations $g_{2i}$ term is used, its expression is

$$g_{2i}(u_i, \overline{M}_i) = \{-\frac{u_i}{f(\overline{M}_i)^2} - \frac{1 - u_i}{(1 - f(\overline{M}_i))^2}\} f'(\overline{M}_i)^2 + \{\frac{u_i}{f(\overline{M}_i)} - \frac{1 - u_i}{1 - f(\overline{M}_i)}\} f''(\overline{M}_i) \tag{70}$$

Thus by (66) we have

$$\left\langle \frac{\partial \widehat{E}}{\partial \beta}\bigg|_{\beta=0} \right\rangle = 0 \tag{71}$$

$$\begin{aligned} \widehat{G}_{1C} \quad = \quad & \sum_{i=1}^{N} [u_i \ln u_i + (1 - u_i) \ln(1 - u_i)] \\ & - \sum_{i=1}^{N} [u_i \ln f(\overline{M}_i) + (1 - u_i) \ln(1 - f(\overline{M}_i))] \\ & - \sum_{k=1}^{C} \sum_{i=1}^{N} \frac{1}{k!} g_{ki}(u_i, \overline{M}_i) \langle X_i^k \rangle \end{aligned} \tag{72}$$

Fixed point equations are obtained by putting $\frac{\partial \widehat{G}_{1C}}{\partial u_i} = 0$, and also noting the point that $\langle X_i \rangle = 0$.

$$u_i = \sigma \left( \left[ \ln \frac{f(\overline{M}_i)}{1 - f(\overline{M}_i)} + \sum_{l=i+1}^{N} \frac{u_l - f(M_l)}{f(M_l)(1 - f(M_l))} f'(M_l) + \sum_{k=2}^{C} \frac{\partial}{\partial u_i} \sum_{i=1}^{N} \frac{1}{k!} g_{ki}(u_i, \overline{M}_i) \langle X_i^k \rangle \right] \right) \tag{73}$$

## Appendix B: Mean-field Expressions for NOISY-OR and SIGMOID BNs

In this section we present formulae for $\widehat{G}_{MC}$,for $C = 1, 2, M = 1, 2$ that was used for the experiments described in Section 4.

$$\widehat{G}_{11} = \sum_{i=1}^{N} (u_i \ln u_i + (1 - u_i) \ln(1 - u_i)) - \sum_{i=1}^{N} (u_i \ln f(\overline{M}_i) + (1 - u_i) \ln(1 - f(\overline{M}_i)) \tag{74}$$

$$\begin{aligned}
\widehat{G}_{12} &= \sum_{i=1}^{N}(u_i \ln u_i + (1-u_i)\ln(1-u_i)) - \sum_{i=1}^{N}(u_i \ln f(\overline{M}_i) + (1-u_i)\ln(1-f(\overline{M}_i)) \\
&\quad - \sum_{i=1}^{N}\left\{\left(-\frac{u_i}{f(\overline{M}_i)^2} - \frac{(1-u_i)}{(1-f(\overline{M}_i))^2}\right) f'(\overline{M}_i)^2 \right. \\
&\quad \left. + \left(\frac{u_i}{f(\overline{M}_i)} - \frac{1-u_i}{1-f(\overline{M}_i)}\right) f''(\overline{M}_i)\right\}\langle X_i^2\rangle \qquad (75)
\end{aligned}$$

where $\langle X_i^2\rangle = \sum_{k=1}^{i-1} w_{ik}^2 u_i(1-u_i)$.

It is a matter of detail to plug in the appropriate functions for a specific network, like the sigmoid or noisy-or. Since it is too cumbersome we do not present explicitly the terms required for $\widehat{G}_{22}$, for a general activation function. Instead we present the expressions used for the sigmoid and noisy-or functions.

For Sigmoid network

$$\begin{aligned}
\widehat{G}_{22} &= \widehat{G}_{12} - 0.5\sum_{i=1}^{N}\left(\sum_{l=1}^{i-1} w_{il}u_l(1-u_l)\left(\overline{M}_l(u_i - \sigma(\overline{M}_l)) + w_{il}u_i(1-u_i)\right)\right. \\
&\quad + \sum_{j=1}^{H+V}\sum_{k=1}^{t-1} w_{jk}w_{ik}u_k(1-u_k)(u_i - \sigma(\overline{M}_i)(u_j - \sigma(\overline{M}_j)) \\
&\quad \left. - \sum_{j=i+1}^{H+V}\sum_{l=i+1}^{H+V}\left(w_{il}(u_l - \sigma(\overline{M}_l) + \overline{M}_i\right)(u_j - \sigma(\overline{M}_j))w_{ji}u_i(1-u_i)\right) \qquad (76)
\end{aligned}$$

For noisy-or network

$$\begin{aligned}
\widehat{G}_{22} &= \widehat{G}_{12} - 0.5\sum_{i=1}^{N}\left(\sum_{j=1}^{i-1} w_{ij}^2 u_j(1-u_j)\left(\frac{u_i(1-u_i)}{f(\overline{M}_i)^2} + \frac{u_i}{f(\overline{M}_i)} - 1)\log(1+e^{\overline{M}_j})\right)\right. \\
&\quad + \sum_{j=1}^{H+V}\sum_{k=1}^{t-1} w_{ik}w_{jk}u_k(1-u_k)(\frac{u_i}{f(\overline{M}_i)} - 1)(\frac{u_j}{f(\overline{M}_j)} - 1) \\
&\quad \left. - \sum_{j=i+1}^{H+V}\left(\sum_{k=i+1}^{H+V} w_{ki}(\frac{u_k}{f(\overline{M}_k)} - 1) + (\frac{u_j}{f(\overline{M}_j)} - 1)\log(1+e^{\overline{M}_i})\right) w_{ji}u_i(1-u_i)\right) \qquad (77)
\end{aligned}$$

In both the expressions (76) and (77) t is chosen to be either $i$ or $j$, whichever is lower; also the activation function, f, in (77) is $f(x) = 1 - e^{-x}$.

The vector $\vec{u}$ is determined by solving fixed point equations obtained by setting

$$\frac{\partial \widehat{G}_{MC}}{\partial u_i} = 0.$$

The gradients required for learning is evaluated by $\frac{\partial \widehat{G}_{MC}}{\partial w_{ij}}$ at the fixed point.